\documentclass[10pt,twocolumn,letterpaper]{article}

\usepackage{iccv}
\usepackage{times}
\usepackage{epsfig}
\usepackage{graphicx}
\usepackage{amsmath}
\usepackage{amssymb}
\usepackage{tabularx}

\usepackage{makecell}

\makeatletter
\newcommand*{\rom}[1]{\expandafter\@slowromancap\romannumeral #1@}
\makeatother
\usepackage[breaklinks=true,bookmarks=false]{hyperref}

\iccvfinalcopy % *** Uncomment this line for the final submission

\def\iccvPaperID{2742} % *** Enter the ICCV Paper ID here	

\begin{document}

%%%%%%%%% TITLE
\title{TAC-GAN -- Text Conditioned Auxiliary Classifier Generative Adversarial Network}

\author{Ayushman Dash$^1$ \qquad John Gamboa$^1$ \qquad Sheraz Ahmed$^3$\\ \qquad Marcus Liwicki$^{14}$ \qquad Muhammad Zeshan Afzal$^{12}$\\
$^1$MindGarage -- University of Kaiserslautern, Germany\\
$^2$Insiders Technologies GmbH, Kaiserslautern, Germany\\
$^3$German Research Center for AI (DFKI),
Kaiserslautern, Germany\\
$^4$University of Fribourg, Switzerland\\
{\tt\small{ayush@rhrk.uni-kl.de, gamboa@rhrk.uni-kl.de, sheraz.ahmed@dfki.de}}\\
 {\tt\small{marcus.liwicki@unifr.ch, afzal@iupr.com}}
%\and
}

\maketitle
%\thispagestyle{empty}

%%%%%%%%% ABSTRACT
\begin{abstract}
In this work, we present the Text Conditioned Auxiliary Classifier Generative Adversarial Network, (TAC-GAN) a text to image Generative Adversarial Network (GAN) for synthesizing images from their text descriptions.
Former approaches have tried to condition the generative process on the textual data; but allying it to the usage of class information, known to diversify the generated samples and improve their structural coherence, has not been explored.
We trained the presented TAC-GAN model on the Oxford-102 dataset of flowers, and evaluated the discriminability of the generated images with Inception-Score, as well as their diversity using the Multi-Scale Structural Similarity Index (MS-SSIM).
Our approach outperforms the state-of-the-art models, i.e., its inception score is $3.45$, corresponding to a relative increase of $7.8\%$ compared to the recently introduced StackGan. A comparison of the mean MS-SSIM scores of the training and generated samples per class shows that our approach is able to generate highly diverse images with an average MS-SSIM of $0.14$ over all generated classes.

\end{abstract}

%%%%%%%%% BODY TEXT
\section{Introduction}

Synthesizing diverse and discriminable images from text is a difficult
task and has been approached from different perspectives.
Making the images realistic, while still capturing the semantics of the text
are some of the challenges that remain to be solved.

% \cite{yan2016attribute2image}
%Realizing a method for such a task has numerous applications, like (1) semi-supervised pre-training for models with limited ground truth, (2) retrieving images given a text query using the generated image as a basis for retrieval, (3) styling images based on textual descriptions, and many more.

Generative Adversarial Networks (GAN) have shown promising results for image
synthesis \cite{goodfellow2014generative}. GANs use an adversarial training
mechanism, based on the minimax algorithm in which a generative model \(G\) and
a discriminative model \(D\) are trained simultaneously with conflicting
objectives. \(G\) is trained to model the data distribution while \(D\) is trained
to classify whether the data is real or generated. The model is then sampled by
passing an input noise vector $\textbf{z}$ through $G$.
The framework has received a lot of attention since its inception (\eg,
\cite{mirza2014conditional, odena2016conditional, reed2016generative, salimans2016improved, zhang2016stackgan}).

Extending these ideas, Odena \etal \cite{odena2016conditional}
proposed the Auxiliary Classifier Generative Adversarial Network
(AC-GAN). In that model, the Generator synthesizes images conditioned on a class
label and the Discriminator not only classifies between real and generated
input images, but also assigns them a class label.

\begin{figure}[b]
\begin{center}
   \includegraphics[width=1.0\linewidth]{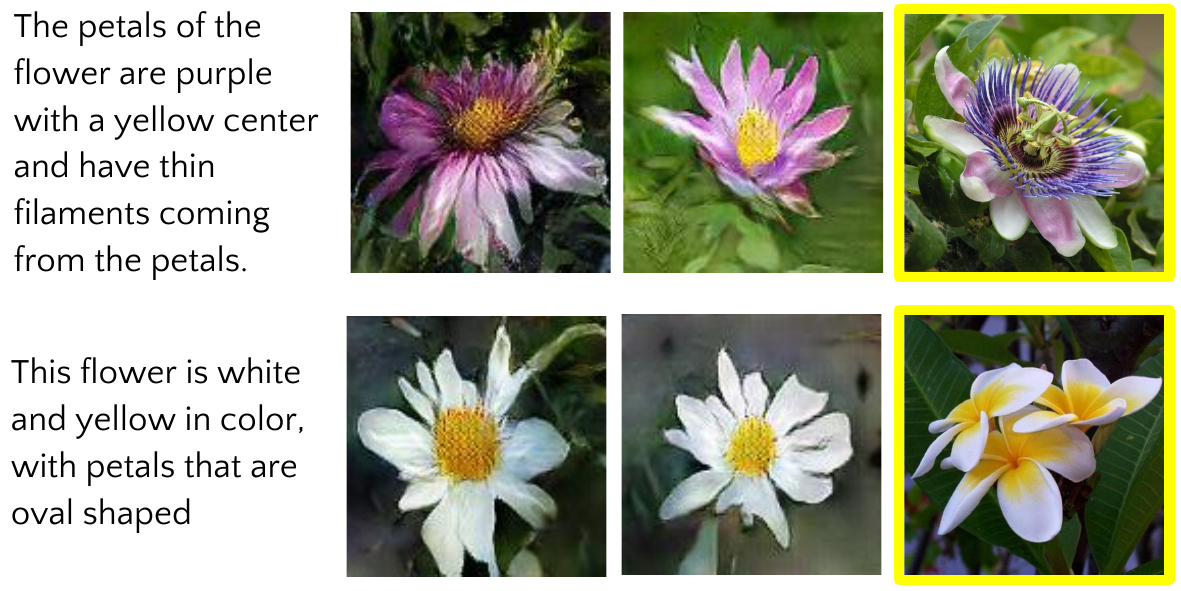}
\end{center}
   \caption{Images generated by the TAC-GAN given a text descriptions. The text on the left were used to generate the images on the right. The highlighted image on the right is a real image corresponding to the text description.} 
\label{fig:frontPage}
\end{figure}

In this paper, we present the Text
Conditioned Auxiliary Classifier Generative Adversarial Network (TAC-GAN), which builds upon the AC-GAN by conditioning the generated images on a text description instead of on
a class label.
In the presented TAC-GAN model, the input vector of
the Generative network is built based on a noise vector $\textbf{z}$ and another vector containing an embedded representation of the
textual description. While the Discriminator is similar to that of
the AC-GAN, it is also augmented to receive the text information as
input before performing its classification.

To evaluate our model, we use the Oxford-102 dataset \cite{nilsback2008automated} of flowers.
Additionally, we use Skip-Thought vectors to generate text embeddings from the image captions. Images generated using \textbf{TAC-GAN} are not only highly discriminable, but are also diverse. 
Similarly to \cite{reed2016generative}, we also show that our model learns to disentangle
the content of the generated images from their style. By interpolating between
different text descriptions, it is possible to synthesize images that differ in content while maintaining the same style.
Evaluation results show that our approach outperforms the state of the art models by $7.8\%$ on inception score. Figure \ref{fig:frontPage} shows some results
generated by our approach.

The rest of the paper is structured as follows. Section~ \ref{sec:relatedWork} provides an overview of the existing methods for text and image generation. The basics of the GAN framework are explained in Section~\ref{sec:gan}, and the architecture of the AC-GAN is presented in Section~\ref{sec:acGan}. Our presented model, TAC-GAN, is described in Section \ref{sec:tacGan}. Section \ref{sec:eval} provides an evaluation and comparison of the presented TAC-GAN with state of the art methods for text to image generation. 
Section \ref{sec:anal} provides an analysis of the images generated using TAC-GAN, and
Section~\ref{sec:conclusion} concludes the paper and provide an overview of the future work.

\begin{figure}[t]
\begin{center}
   \includegraphics[width=1.0\linewidth]{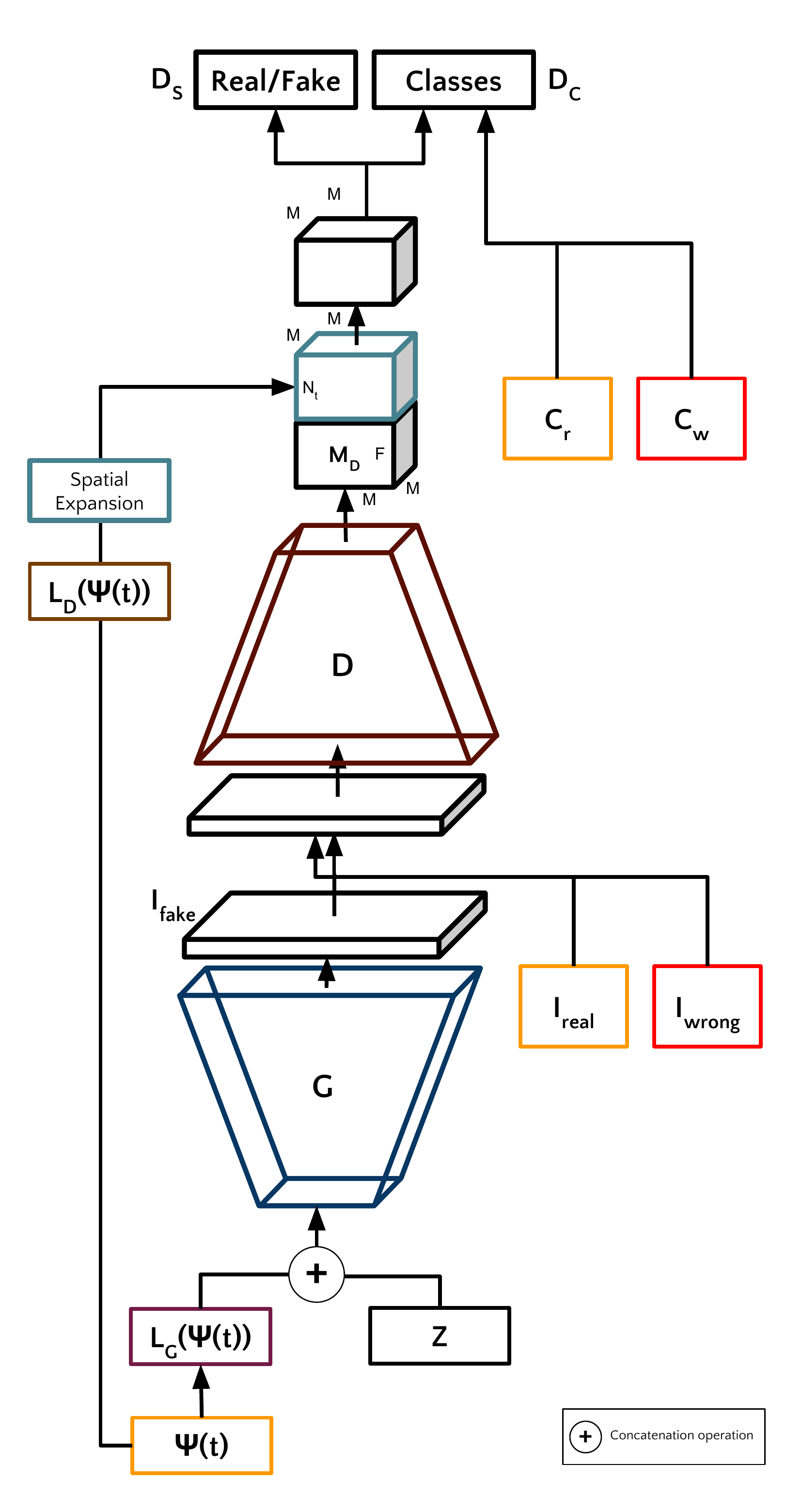}
\end{center}
   \caption{The architecture of the TAC-GAN. Here, $t$ is a text description of an image, $z$ is a noise vector of size $N_z$, $I_{real}$ and $I_{wrong}$ are the real and wrong images respectively, $I_{fake}$ is the image synthesized by the generator network $G$, $\Psi(t)$ is the text embedding for the text $t$ of size $N_t$, and $C_r$ and $C_w$ are one-hot encoded class labels of the $I_{real}$ and $I_{wrong}$, respectively. $L_G$ and $L_D$ are two neural networks that generate latent representations of size $N_l$ each, for the text embedding $\Psi(t)$. $D_S$ and $D_C$ are the probability distribution that the Discriminator outputs over the sources (real/fake) and the classes respectively.}
\label{fig:tacGan}
\end{figure}

\begin{figure*}[t]
\begin{center}
   \includegraphics[width=1.0\linewidth]{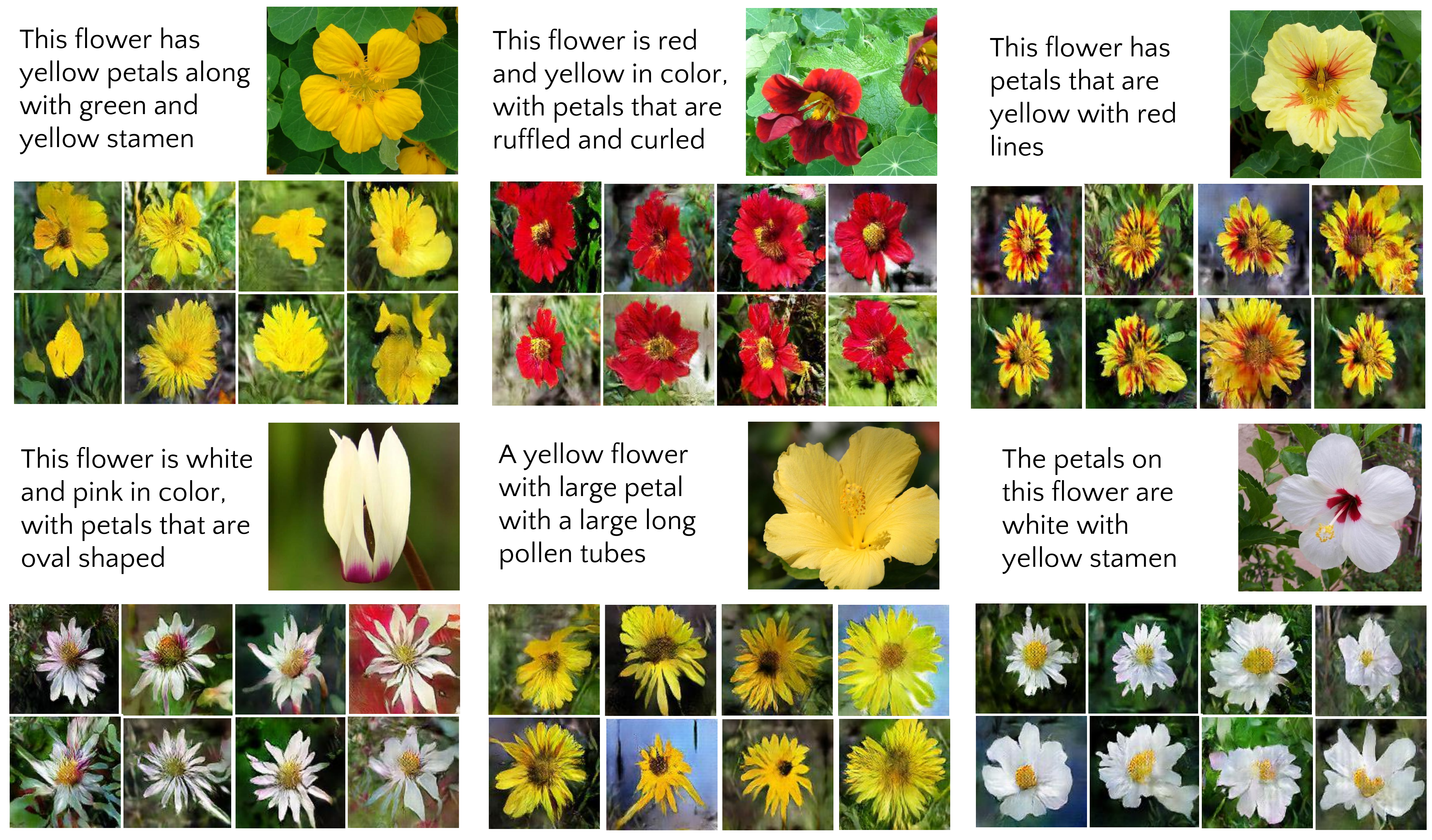}
\end{center}
   \caption{Images synthesized from text descriptions using different noise vectors. In each block, the images at the bottom are generated from the text embeddings of the image description and a noise vector. The image on the top of each block are real images corresponding to the text description.}
\label{fig:bigResults}
\end{figure*}

\section{Related Work}
\label{sec:relatedWork}

A considerable amount of work on
image modelling
focused on other tasks such as image inpainting (\eg,
\cite{hays2007scene}) or texture generation (\eg,
\cite{efros1999texture}). These tasks produced new images
based on an existing image. However, image synthesis not
relying on such existing information has not had much
success until recently \cite{radford2015unsupervised}.

In the last few years, approaches based on Deep Learning have arised,
which are able to synthesize images with varied success.
Variational Autoencoders (VAE) \cite{doersch2016tutorial, kingma2013auto},
once trained, can be interpreted as generative models that produce
samples from a distribution that approximates that of the training set.
The DRAW model \cite{gregor2015draw} extends the VAE architecture
by using recurrent networks and an attention mechanism,
transforming it into a sequence
encoding/decoding process such that ``the network decides at
each time-step `where to read' and `where to write' as well as
`what to write'", and being able to produce highly realistic
handwritten digits. Mansimov \etal \cite{mansimov2015generating},
extended the DRAW model by using a Bidirection RNN and conditioning
the generating process on text captions, producing results that were
slightly better than those of other state of the art models.

Alternatively, Generative
Adversarial Networks (GAN) \cite{goodfellow2014generative} have gained
a considerable amount of interest since its inception, having been
used in tasks such as
single-image super resolution \cite{ledig2016photo},
Simulated+Unsupervised learning \cite{shrivastava2016learning},
image-to-image translation \cite{isola2016image} and
semantic image inpainting \cite{yeh2016semantic}.
The basic framework, however, tends to produce blurry images, and the
quality of the generated images tends to decrease as the resolution
increases.
Some approaches to tackle these problems have focused on iteratively
refining the generated images. For example, in the Style and Structure GAN
(S$^2$-GAN) model 
\cite{wang2016generative}, a first GAN is used to produce the image
structure, which is then fed into a second GAN, responsible for the image
style. Similarly, the Laplacian GANs (LAPGAN) \cite{denton2015deep}
can have an indefinite number of stages, integrating ``a conditional
form of GAN model into the framework of a Laplacian pyramid".
Other attempts to solve the problems of the basic framework have tried
making the network better aware of
the data distribution that the GAN is supposed to model.
By conditioning the input on specific class labels,
the Conditional GAN (CGAN) \cite{mirza2014conditional} was capable of
producing higher resolution images.
Alternatively, the Auxiliary Classifier GAN (AC-GAN) \cite{odena2016conditional}
has been shown to be capable of synthesizing structurally coherent
$128 \times 128$ images by
training the discriminator to also classify its input.

Of special relevance to this work is the conditioning of the generative
process additionally on text. Reed \etal
\cite{reed2016generative}, following the work on the
Generative Adversarial What-Where Networks \cite{reed2016learning},
were able to make the synthesized images correspond to a textual
description used as input, with a resolution of $64 \times 64$. With
a similar approach, the StackGAN model \cite{zhang2016stackgan}
leverages the benefits of using multiple stages and is capable of generating
highly realistic $256 \times 256$ images.

\section{Background}

This section describes some of the existing work our approach
is built upon. Section \ref{sec:gan} describes the basic GAN framework,
and Section \ref{sec:acGan} describes the AC-GAN.

\subsection{Generative Adversarial Networks}
\label{sec:gan}

The basic framework of Generative Adversarial Networks (GAN) was first introduced by Goodfellow \etal \cite{goodfellow2014generative}. % The following formulation is based on that of \cite{odena2016conditional}.
GANs are generative models that introduce a new paradigm of adversarial training \cite{goodfellow2014explaining} in which a generative model \(G\) and a discriminative model \(D\) are trained simultaneously. \(G\) is trained to model the data distribution, while \(D\) is trained to classify if the data is real or generated.
In this case both \(G\) and \(D\) are both neural networks.
%to learn the data distribution and generate samples that realistically close to the data in the data distribution. 
%A GAN is composed of two networks: a discriminative network $D$ and a
%generative network $G$. 
Let $\mathcal{X}$ be a dataset used for training the
GAN, and $I_{real}$ denote a sample from $\mathcal{X}$.
The two networks are then trained to minimize conflicting objectives.
Given an input noise vector $\bf{z}$,
$G$ is trained to generate samples $I_{fake}$ that are similar to those of the
training set $\mathcal{X}$, \ie, to generate fake samples.
$D$, on the other hand, receives a sample $I$ as input and is trained to return
a probability distribution $P(S | I)$, interpreted as the probability that
$I$ pertains to the dataset $\mathcal{X}$, (\ie, is a real sample).

Specifically, $D$ is trained to maximize, and $G$ is trained to minimize,
the following value:
\begin{equation}
\mathbb{E}[\log P(S = real | I_{real})] + \mathbb{E}[\log P(S = fake | I_{fake})]
\end{equation}

In the context of this model, $\mathcal{X}$ is composed of images,
and each $I$ is an image.

\begin{figure*}
\begin{center}
   \includegraphics[width=1.0\linewidth]{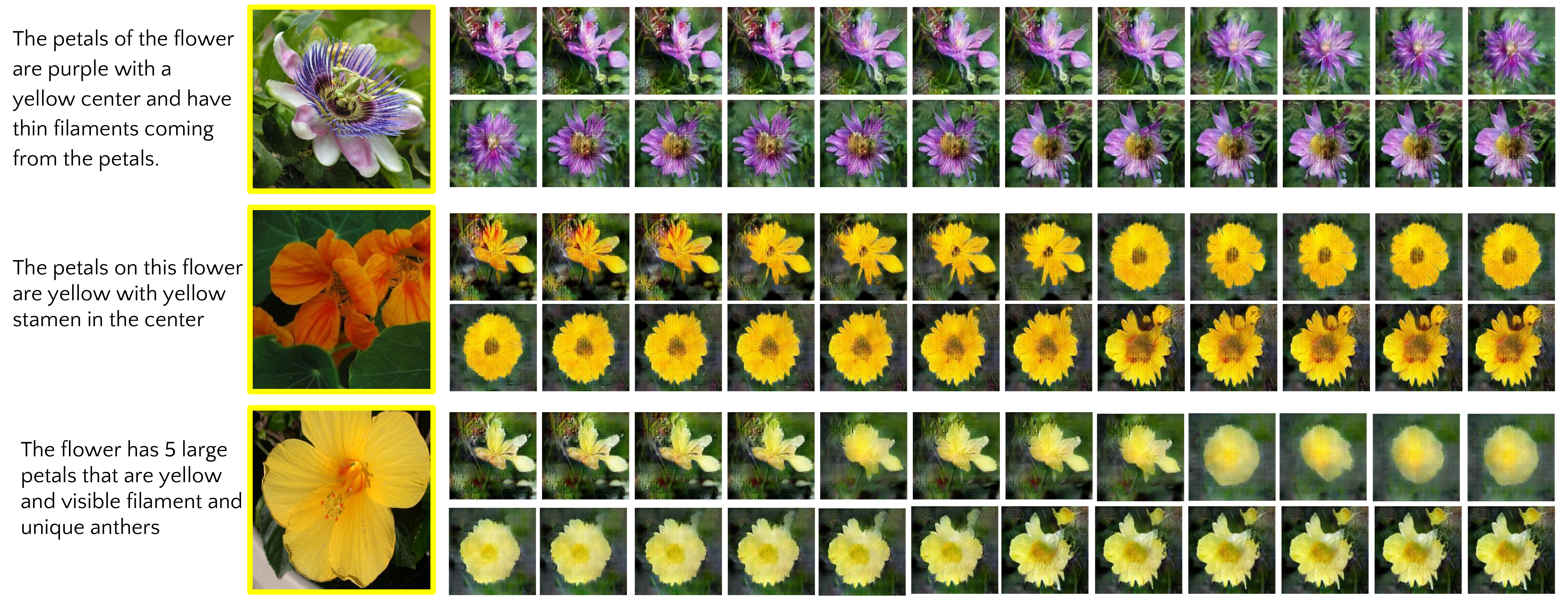}
\end{center}
   \caption{For each block, two noise vectors $\textbf{z}_1$ and $\textbf{z}_2$ are generated. They are used to synthesize the images in the extremes. For the images in between, an interpolation between the two vectors is used. The text embedding used to produce the images is the same for the entire block. It is produced from the textual description in the left. For comparison, the Ground Truth image is highlighted. As can be seen, the style of the synthesized images changes, but the content remains roughly the same, based on that of the text input.}
\label{fig:z-interpolation}
\end{figure*}

\subsection{Auxiliary Classifier Generative Adversarial Networks (AC-GAN)}
\label{sec:acGan}
The AC-GAN \cite{odena2016conditional} is a variant of the GAN architecture in which \(G\) conditions the generated data on its class label, and the Discriminator performs an auxiliary task of classifying the synthesized and the real data into their respective class labels. In this setting every produced image is associated with a class label $c$ and a noise vector $\textbf{z}$, which are used by \(G\)  to generate images $I_{fake} = G(c,\textbf{z})$. The Discriminator of an AC-GAN outputs a probability distribution over sources (fake or real), as well as a probability distribution over the class labels: $D_S(I) = P(S \mid I)$ and $D_C(I) = P(C \mid I)$. The objective function consists of two parts: (1) the log-likelihood of the correct source $L_S$;
and (2) the log-likelihood of the correct class $L_C$:
\begin{equation}
\begin{split}
L_S = ~~&\mathbb{E}[\log P(S = real ~~|~ X_{real} )] ~+ \\
        &\mathbb{E}[\log P(S = fake ~|~ X_{fake})] \\
\end{split}
\end{equation}
\begin{equation}
\begin{split}
L_C = ~~&\mathbb{E}[\log P(C = c ~~|~ X_{real} )] ~+ \\
        &\mathbb{E}[\log P(C = c ~|~ X_{fake})]
\end{split}
\end{equation}

During training, $D$ maximizes $L_C + L_S$, while $G$ minimizes $L_C - L_S$.

\section{Text Conditioned Auxiliary Classifier Generative Adversarial Network (TAC-GAN)}
\label{sec:tacGan}

Our proposed model, the TAC-GAN, generates images of size $128 \times 128$ that
comply to the content of the input text.

%The Generator of the TAC-GAN differs from the AC-GAN in the following way. Let $t$ denote the text description of the image and $\Psi$ be a text embedding function. Every produced image has a corresponding text embedding $\Psi(t)$ and a noise vector $\textbf{z}$. $\Psi(t)$ and $\textbf{z}$ are used to generate a new noise vector $\hat{\textbf{z}}_c$, which is used by \(G\) to generate $I_{fake}$.

%The Discriminator of the TAC-GAN is similar to that of the AC-GAN. However, to model
%the textual information, it uses a procedure similar to that of
%\cite{reed2016generative}

To train our model, we use the Oxford-102 flowers dataset, which has, for every image, a class label and at least five text descriptions.
For implementing TAC-GAN we use the Tensorflow \cite{abadi2016tensorflow} implementation of a Deep Convolutional Generative Adversarial Network (DCGAN)\footnote{\url{https://github.com/carpedm20/DCGAN-tensorflow}} \cite{radford2015unsupervised}, in which \(G\) is modeled as a Deconvolutional (fractionally-strided convolutions) Neural Network, and \(D\) is modeled as a Convolutional Neural Network (CNN). For our text embedding function $\Psi$, we use Skip-Thought vectors to generate an embedding vector of size $N_t$.

We start by describing the model
architecture, and proceed with the training procedure.

\subsection{Model Architecture}
\label{sec:modelArchitecture}

In our approach we introduce a variant of AC-GAN, called Text Conditioned Auxiliary Classifier Generative Adversarial Networks (TAC-GAN), to synthesize images from text. As opposed to AC-GANs, we condition the generated images from TAC-GANs on text embeddings and not on class labels.

Figure \ref{fig:tacGan} shows the architecture of a TAC-GAN.

\subsubsection{Preliminaries}

Let $\mathcal{X}=\{\mathcal{X}_i \mid i = 1,...,n\}$ be a dataset, where $\mathcal{X}_i$ are the data instances and $n$ is the number of instances in the dataset ($|\mathcal{X}|=n$). Every data instance is a tuple $\mathcal{X}_i = (I_i, T_i, C_i)$, where $I_i$ is an image, $T_i=(t_i^1, t_i^2,..., t_i^k)$ is a set of $k$ text descriptions of the image, and $C_i$ is the class label to which the image corresponds.
For training the TAC-GAN we randomly pick a text description $t_i^j$ from $T_i$ and generate a text embedding $\Psi(t_i^j) \in \mathbb{R}^{N_t}$. We then generate a latent representation $l_i^j = \mathcal{L}_G(\Psi(t_i^j))$ of the text embedding, where $\mathcal{L}_G$ is a fully connected neural network with $N_l$ neurons. Therefore $l_i^j \in \mathbb{R}^{N_l}$, where $N_l$ is a hyperparameter of the model. This representation is then concatenated to a noise vector $\textbf{z} \in [-1,1]^{N_z}$, creating a vector $\textbf{z}_c \in \mathbb{R}^{N_l+N_z}$. Here, $N_z$ is also a hyperparameter of the model.
$\textbf{z}_c$ is then passed through a fully connected layer $FC_G$ with $8 * 8 * (8 * N_c)$ neurons, where $N_c$ is another hyperparameter of the model. The output of $FC_G$ is finally reshaped into a convolutional representation $\hat{\textbf{z}_c}$ of shape $8 \times 8 \times (8 * N_c)$.

\subsubsection{Generator Network}

The Generator Network is very similar to that of the AC-GAN. However, instead of
feeding the class label to which the synthesized image is supposed to pertain, we
input the noise vector $\hat{\textbf{z}_c}$, containing information related to
the textual description of the image.

In our model, $G$ is a neural network consisting of a sequence of transposed convolutional layers. It outputs an up-scaled image $I_f$ (fake image) of shape $128 \times 128 \times 3$.

\subsubsection{Discriminator Network}

Let $I_r$ denote the real image from the dataset corresponding to the text description $t_i^j$, and $I_w$ denote another image that does not correspond to $t_i^j$. Additionally, let $C_r$ and $C_w$ correspond to the class labels of $I_r$ and $I_w$, respectively.
We use $I_w$ to teach the Discriminator to
consider fake any image that does not belong to the desired class.
%For $I_w$ a random text description $t_w$ is selected.
In the context of the Discriminator, let $t_r = t_i^j$ be the text description of the real and fake images (notice that the fake image was generated based on the text description of $t_r$). A new latent representation $l_r = \mathcal{L}_D(\Psi(t_r))$ for the text embeddings is generated, where $\mathcal{L}_D$ is another fully connected neural network with $N_l$ neurons, and hence $l_r \in \mathbb{R}^{N_l}$.
%Along with this $C_r$, the class label corresponding to the $I_r$, and $C_w$, the class label corresponding to the $I_w$, are extracted too. A random text description $I_r$ is generated.

A set $\mathcal{A} = \{(I_f, C_f, l_f), (I_r, C_r, l_r), (I_w, C_w, l_w)\}$ is created, to be used by the discriminator. The set is composed of three tuples containing an image, a corresponding class label, and a corresponding text embedding.
Notice that $I_f$ was created conditioned on the same class as $C_r$ and on the same text description as $l_r$. Therefore, $C_f = C_r$ and $l_f = l_r$.
Similarly, $I_w$ is \textit{wrong} because
it does not correspond to its $l_w$, since $l_w = l_r$.
Therefore, it is possible to convert the previous definition of $\mathcal{A}$ into
$\mathcal{A} = \{(I_f, C_r, l_r), (I_r, C_r, l_r), (I_w, C_w, l_r)\}$. The
values of these images are used by the discriminator in the following way.

The Discriminator network is composed by a series of convolutional layers
and receives an image $I$ (any of the images from $\mathcal{A}$). By passing
throught the convolutional layers, the image is downsampled into an
image $\bf{M_D}$ of size $M \times M \times F$, where
$M$ and $F$ are hyperparameters of the model.
$l_r$ is replicated spatially to form a vector of shape $M \times M \times N_l$, and is concatenated with $\bf{M_D}$ in the $F$ (channels) dimension, similarly to
what is done in \cite{reed2016generative}.
This concatenated vector is then fed to another convolutional layer with spatial dimension $M \times M$. Finally, two fully connected layers $FC_1$ and $FC_2$ are used with $1$ and $N_c$ neurons, respectively, along with the sigmoid activation function. $FC_1$ produces a probability distribution $D_S$ over the sources (real/fake), while $FC_2$ produces a probability distribution $D_C$ over the class labels. Details of the architecture are described in Figure \ref{fig:tacGan}.

This discriminator design was inspired by the GAN-CLS model proposed by Reed \etal \cite{reed2016generative}. The parameters of $\mathcal{L}_G$ and $\mathcal{L}_D$ are trained along with the TAC-GAN. The training process is described in detail in Section \ref{sec:training}.

\begin{table}
\caption{Global parameters used in the model}
\label{tab:globalParameters}
\begin{center}
\begin{tabular}{c|r||c|r}
\textbf{Parameter} & \textbf{Value} & \textbf{Parameter} & \textbf{Value} \\
\hline
Learning Rate & $0.0002$  & $N_t$ & 4800 \\
$\beta1$      & $0.5$     & $N_l$ & 100  \\
$\beta2$      & $0.999$   & $N_z$ & 100  \\
$\epsilon$    & $10^{-8}$ & $N_c$ & 64   \\
$M$           & $8$       & $N_f$ & 64   \\
$F$           & $384$     &       &
\end{tabular}
\end{center}
\end{table}

\begin{figure*}
\begin{center}
   \includegraphics[width=1.0\linewidth]{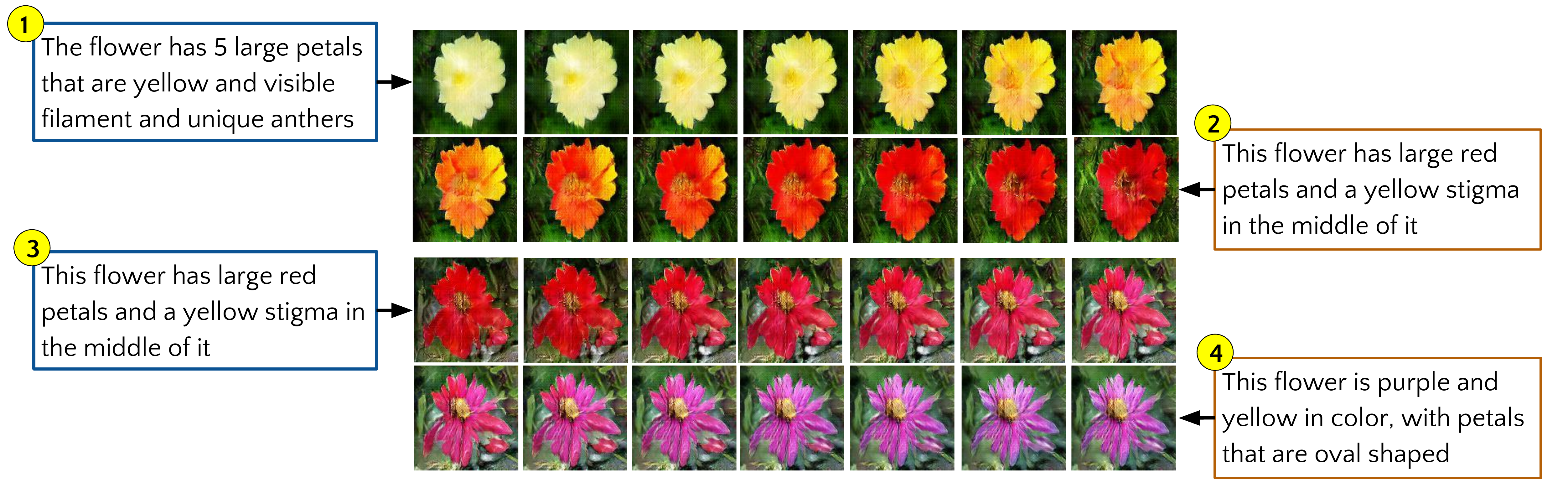}
\end{center}
   \caption{All images generated in the first and second row use the same noise vector $\textbf{z}_1$. Similarly, all images generated in the third and fourth rows use the same noise vector $\textbf{z}_2$. The first image of the first row and the last image of the second row use a text embedding that was constructed from the captions 1 and 2, respectively. An interpolation between these two embeddings was used for synthesizing all images in between them. The first image of the third row and the last image of the fourth row use a text embedding constructed from the captions 3 and 4. An interpolation between these two embeddings was used for synthesizing all images in between them. Notice that captions 2 and 3 are the same, but we use a different noise vector to generate the two different outputs.}
   
   %All images generated in each row use the same noise vector. Two text embeddings for the descriptions in the extremes are computed and used to generate the images in the extremes. The images in between use an interpolated text embedding computed as a weighted average between them. As can be seen, while the style remains roughly the same, the content changes according to the text input.}
\label{fig:t-interpolation}
\end{figure*}

\subsection{Implementation Details}
\label{sec:implementationDetails}

Our Generator network is composed by three transposed
convolutional layers with 256, 128 and 64 filter maps, respectively.
The output of each layer has a size twice as big as that of the images
fed to them as input.
The output of the last layer is the produced $I_f$ used as input to the
Discriminator.
$D$ is composed of three convolutional layers with 128, 256, and 384
filter maps, respectively. The output of the last layer is $\bf{M_D}$.
The kernel size of all the convolutional layers until the generation of
$\bf{M_D}$ is $5 \times 5$.

After the concatenation between $\bf{M_D}$ and the spatially replicated
$l_r$, the last convolutional layer is composed of 512 filter maps of
size $1 \times 1$ and stride 1.

We always use \textit{same} convolutions, and training is performed using
the Adam optimizer \cite{kingma2014adam}.
Table \ref{tab:globalParameters} shows the hyperparameters used
by our model.

\subsection{Training}
\label{sec:training}
%We show how our model could be extended to use any other type of information that
%might help the quality of the synthesized images.

%As explained in Section \ref{sec:tacGan}, the Generator network of a TAC-GAN
%is conditioned on textual information.
In this section, we explain how training is performed in the TAC-GAN model.
We then explain how the loss function could be easily extended so that any other
potentially useful type of information can be leveraged by the model.

\subsubsection{Training Objectives}

\begin{figure*}[t]
\begin{center}
   \includegraphics[width=0.8\linewidth]{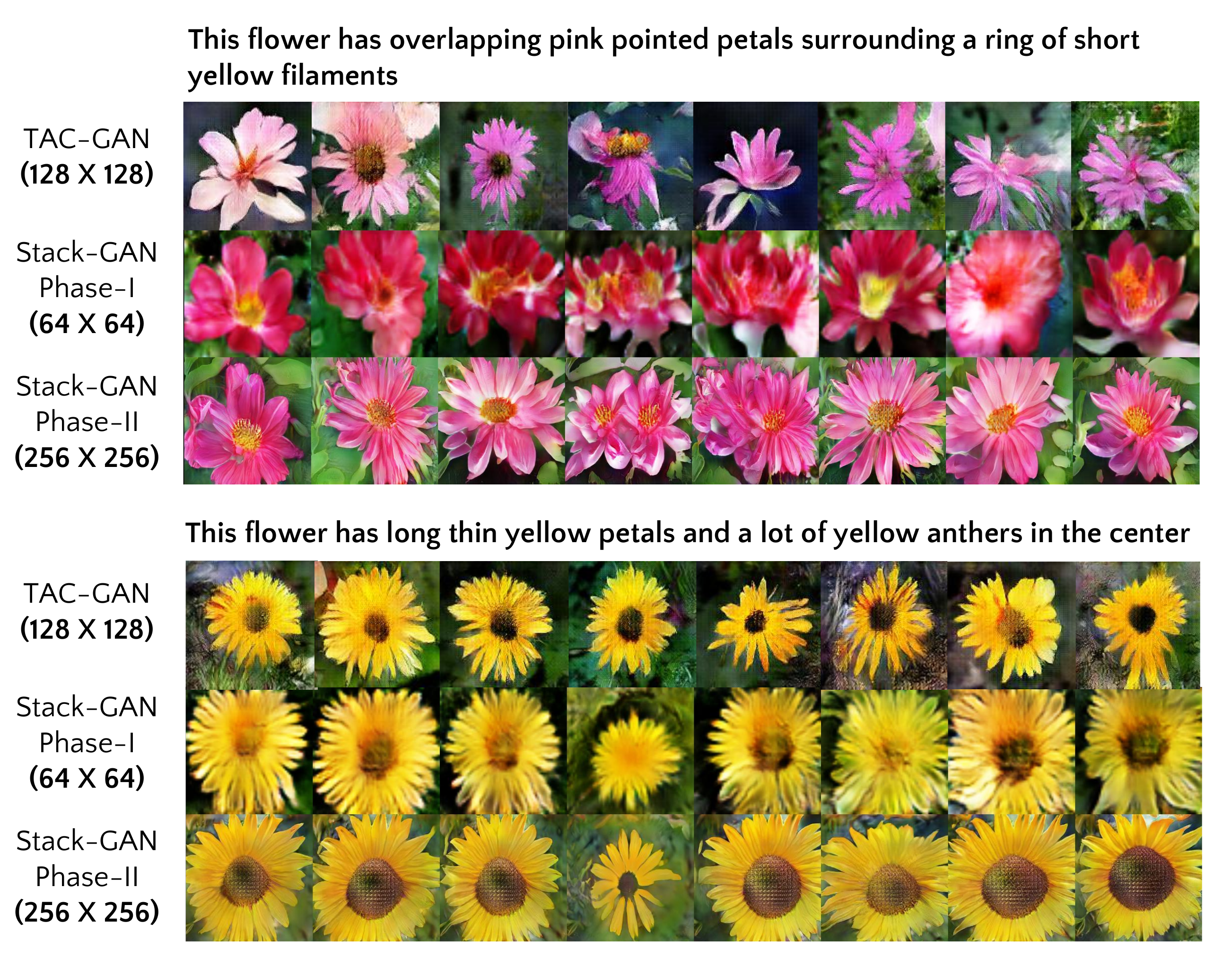}
\end{center}
   \caption{A comparison between the results of our model (TAC-GAN) and those the StackGAN \cite{zhang2016stackgan} in Phase-\rom{1} and Phase-\rom{2}. The images were synthesized based on the caption on the top of each block.}
\label{fig:comparison}
\end{figure*}

Let $L_{D_S}$ denote the training
loss related to the source of the input (real, fake or wrong). Then $L_{D_S}$ is calculated as a sum of the binary cross entropy,
denoted by $H$ below, between
the output of the discriminator and the desired value for each of the images:

\begin{equation}
\begin{split}
L_{D_S} = H\Big(D_s(I_r, l_r), 1\Big) &+ H\Big(D_s(I_f, l_r), 0\Big) \\
                                 &+ H\Big(D_s(I_w, l_r), 0\Big)
\end{split}
\end{equation}

Similarly, let $L_{D_C}$ denote the training loss related to the class to which
the input image is supposed to pertain:

\begin{equation}
\begin{split}
L_{D_C} = H\Big(D_c(I_r, l_r), C_r\Big) &+ H\Big(D_c(I_f, l_r), C_r\Big)\\
                                       &+ H\Big(D_c(I_w, l_r), C_w\Big)
\end{split}
\end{equation}

The Discriminator then minimizes the training objective $L_{D_C} + L_{D_S}$.

In the case of the Generator network, there are no real or wrong images to
be fed as input. The training loss, given by $L_{G_C} + L_{G_S}$, can
therefore be defined in terms of only
the Discriminator's output for the generated image ($L_{G_S}$), and the expected
class to which the synthesized image is expected to pertain ($L_{G_C}$):

\begin{equation}
L_{G_S} = H\Big(D_s(I_f, l_r), 1\Big)
\end{equation}

\begin{equation}
L_{G_C} = H\Big(D_c(I_f, l_r), C_r\Big)
\end{equation}

Notice that, while $L_{D_S}$ penalizes the cross-entropy between $D_S(I_f, l_r)$ and
0 (making $D$ better at deeming generated images fake), $L_{G_S}$ penalizes the
cross-entropy between $D_S(I_f, l_r)$ and 1 (approximating the distribution of the
generated images to that of the training data).

\subsubsection{Extending the current model}

The training losses for both the $G$ and $D$ is a sum of the losses
corresponding to different types of information. In the presence of other types
of information, one can easily extend such a loss by adding a new factor
to the sum.

Specifically, assume a new dataset $\mathcal{Y}$ is present,
containing, for each real image $I_r$ in $\mathcal{X}$, a corresponding
vector $Q_r$ containing details, such as the presence of certain objects in the
$I_r$, the location where such details appear, \etc. The Discriminator could be
extended to output the probability distribution $D_{\mathcal{Y}}(I)$ from any
input image. Let $L_{D_{\mathcal{Y}}}$ denote the related Discriminator loss:

\begin{equation}
\begin{split}
L_{D_{\mathcal{Y}}} = H\Big( D_{\mathcal{Y}}(I_r, l_r), Q_r \Big) &+ H\Big(D_{\mathcal{Y}}(I_f, l_r), Q_r \Big)\\
                                                             &+ H\Big(D_{\mathcal{Y}}(I_w, l_r), Q_w \Big)
\end{split}
\end{equation}

And $L_{G_{\mathcal{Y}}}$ denote the related loss for the Generator:

\begin{equation}
L_{G_{\mathcal{Y}}} = H\Big( D_{\mathcal{Y}}(I_f, l_r) , Q_f \Big)
\end{equation}

Training could then be performed by adding $L_{D_{\mathcal{Y}}}$ to the
Discriminator's loss, and $L_{G_{\mathcal{Y}}}$ to the Generator's loss. This idea is a generalization of extensions of the GAN framework that have been proposed in previous works
(\eg, \cite{odena2016conditional} and \cite{reed2016generative}).

\section{Evaluation}
\label{sec:eval}

\begin{table}[t]
\caption{Inception Score of the generated samples on the Oxford-102 dataset.}
\label{tab:inceptionScore}
\begin{center}
\begin{tabular}{l|c}
\textbf{Model} & \textbf{Inception Score}\\
\hline
\textbf{TAC-GAN} & \textbf{$3.45 \pm 0.05$} \\
StackGan     &  $3.20 \pm .01$     \\
GAN-INT-CLS  &  $2.66 \pm .03$
\end{tabular}
\end{center}
\end{table}

Figure \ref{fig:bigResults} shows some of the generated images for a given
text description, along with the ground truth image from the Oxford-102 dataset.
It can be seen that our approach generates results whose content is in
accordance to the input text. In Figure \ref{fig:comparison}, we compare the
results of our approach
to those of the StackGAN model \cite{reed2016generative}.

Evaluating generative models, like GANs, have always been a challenge
\cite{mansimov2015generating}.
While no standard evaluation metric exists, recent works have introduced a lot of new metrics \cite{goodfellow2014generative,odena2016conditional,radford2015unsupervised}. However, not all of these metrics are useful for evaluating GANs \cite{theis2015note}. Salimans \etal \cite{salimans2016improved} proposed that inception score can be considered as a good evaluation metric for GANs to show the discriminability of the generated images. Odena \etal \cite{odena2016conditional} have shown that Multi-Scale Structural Similarity (MS-SSIM) can be used to measure the diversity of the generated images.

We use inception score to evaluate our approach.
Table \ref{tab:inceptionScore} shows the inception score of the
images generated by our model compared to
those of similar models for image generation from text. It can be seen
that our approach produces better scores than those of other state of the
art approaches. This shows that our approach generates images with higher discriminabality.

\begin{figure}[t]
\begin{center}
   \includegraphics[width=1.0\linewidth]{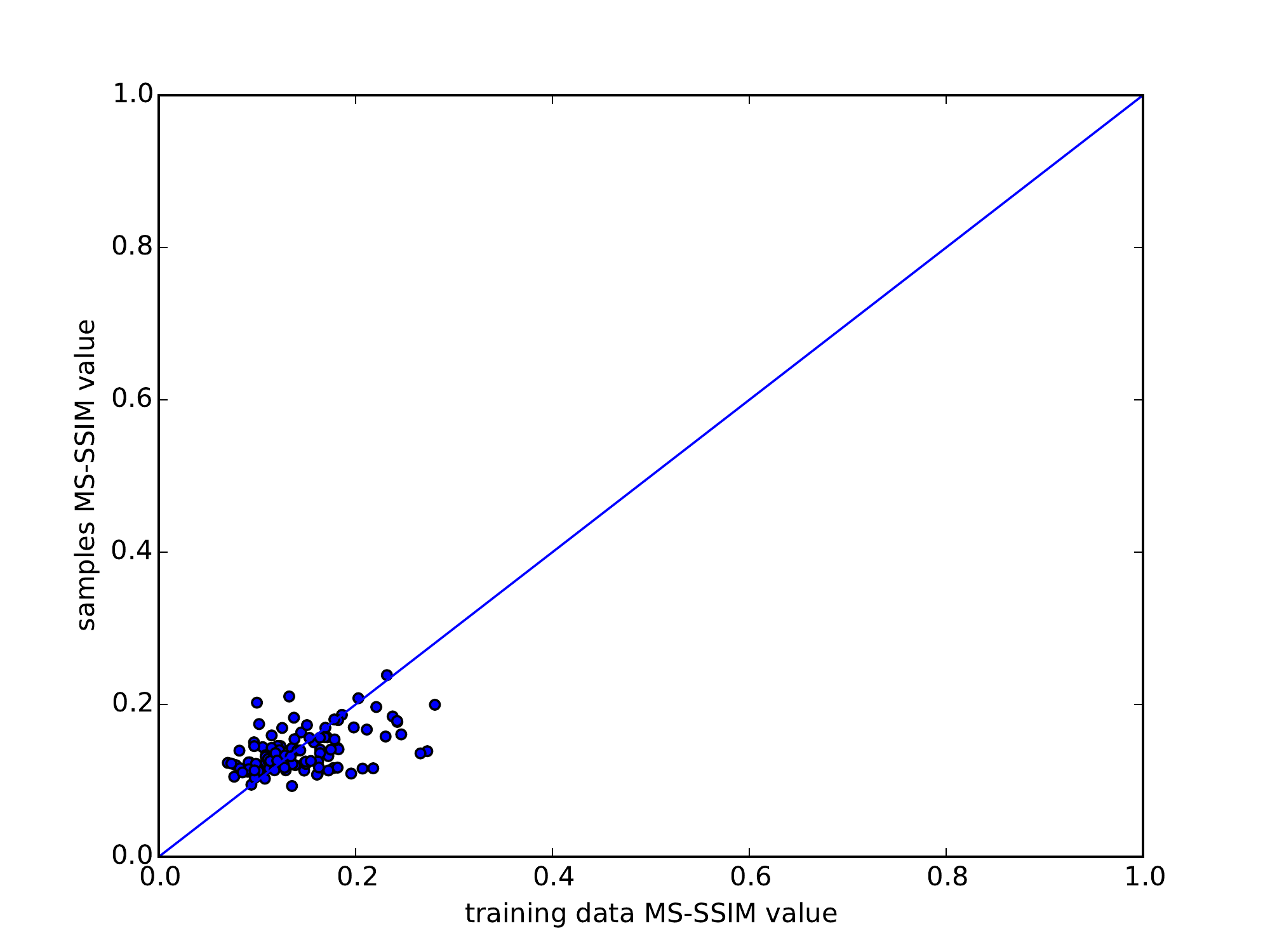}
\end{center}
   \caption{Comparison of the mean MS-SSIM scores of the images of the training data in the Oxford-102 flowers dataset and the sample data generated by the TAC-GAN. Each point represents a class and denotes how similar to each other the images of that class are in the two data sets. The maximum score for samples of a class in the training set is $0.28$ and the mean of the average MS-SSIM over all the classes is $0.14 \pm 0.019$. The maximum MS-SSIM score for samples of a class generated by the TAC-GAN is $0.23$ and the mean of the average MS-SSIM over all classes for the generated images is $0.13 \pm 0.016$.}
\label{fig:ms-ssim}
\end{figure}

To show that our model produces very diverse sample, we used
the MS-SSIM \cite{ma2016group, wang2004image}. Figure \ref{fig:ms-ssim} shows the mean MS-SSIM values of each class of the training dataset, compared to that
of each class of the sampled images. It can be seen that our model
produces more diverse images than those of the training data,
which corroborates with the results from
\cite{odena2016conditional}.

\section{Analysis and Discussion}
\label{sec:anal}
Additionally, we show that our model confirms findings in other approaches
(e.g., \cite{reed2016generative, zhang2016stackgan}), i.e., it
learns separate representations for the style and the content of the generated
images. The images in Figure \ref{fig:z-interpolation} are produced by
interpolating between two different noise vectors, while maintaining
the same input text. While the content of
the image remains roughly unchanged, its style transitions smoothly
from one of the vectors to the other.

Similarly, because we use vector embeddings for the textual descriptions,
it is possible to interpolate between the two embeddings. In Figure
\ref{fig:t-interpolation}, we fix the same $\textbf{z}$ for all generated
images and interpolate between the vector embeddings resulting from applying
$\Psi$ to two text descriptions. It can be seen that the resulting images
maintain a similar style while smoothly transitioning from one content to
another.

\section{Conclusion and Future Work}
\label{sec:conclusion}

We described the TAC-GAN, a model capable of generating images based on textual descriptions.
The results produced by our approach are slightly better to those of
other state of the art approaches.

The model is easily extensible: it is possible to condition the
networks not only on text, but in any other type of potentially useful
information.
It remains to be examined what influence the usage of other types of information
might have in the stability of training, and how much they help, as opposed to
hinder, the capacity of the model in producing better quality, higher resolution
images.

Many approaches have used a multi-staged architecture, where images produced
in the first phase are iteratively refined in subsequent phases. We believe
that the results of our model can benefit from such a pipeline, and be able to
further improve the results reported in this work.

{\small
\bibliographystyle{ieee}
\bibliography{egbib}

\begin{thebibliography}{10}\itemsep=-1pt

\bibitem{abadi2016tensorflow}
M.~Abadi, A.~Agarwal, P.~Barham, E.~Brevdo, Z.~Chen, C.~Citro, G.~S. Corrado,
  A.~Davis, J.~Dean, M.~Devin, et~al.
\newblock Tensorflow: Large-scale machine learning on heterogeneous distributed
  systems.
\newblock {\em arXiv preprint arXiv:1603.04467}, 2016.

\bibitem{denton2015deep}
E.~L. Denton, S.~Chintala, R.~Fergus, et~al.
\newblock Deep generative image models using a laplacian pyramid of adversarial
  networks.
\newblock In {\em Advances in neural information processing systems}, pages
  1486--1494, 2015.

\bibitem{doersch2016tutorial}
C.~Doersch.
\newblock Tutorial on variational autoencoders.
\newblock {\em arXiv preprint arXiv:1606.05908}, 2016.

\bibitem{efros1999texture}
A.~A. Efros and T.~K. Leung.
\newblock Texture synthesis by non-parametric sampling.
\newblock In {\em Computer Vision, 1999. The Proceedings of the Seventh IEEE
  International Conference on}, volume~2, pages 1033--1038. IEEE, 1999.

\bibitem{goodfellow2014generative}
I.~Goodfellow, J.~Pouget-Abadie, M.~Mirza, B.~Xu, D.~Warde-Farley, S.~Ozair,
  A.~Courville, and Y.~Bengio.
\newblock Generative adversarial nets.
\newblock In {\em Advances in neural information processing systems}, pages
  2672--2680, 2014.

\bibitem{goodfellow2014explaining}
I.~J. Goodfellow, J.~Shlens, and C.~Szegedy.
\newblock Explaining and harnessing adversarial examples.
\newblock {\em arXiv preprint arXiv:1412.6572}, 2014.

\bibitem{gregor2015draw}
K.~Gregor, I.~Danihelka, A.~Graves, D.~J. Rezende, and D.~Wierstra.
\newblock Draw: A recurrent neural network for image generation.
\newblock {\em arXiv preprint arXiv:1502.04623}, 2015.

\bibitem{hays2007scene}
J.~Hays and A.~A. Efros.
\newblock Scene completion using millions of photographs.
\newblock In {\em ACM Transactions on Graphics (TOG)}, volume~26, page~4. ACM,
  2007.

\bibitem{isola2016image}
P.~Isola, J.-Y. Zhu, T.~Zhou, and A.~A. Efros.
\newblock Image-to-image translation with conditional adversarial networks.
\newblock {\em arXiv preprint arXiv:1611.07004}, 2016.

\bibitem{kingma2014adam}
D.~Kingma and J.~Ba.
\newblock Adam: A method for stochastic optimization.
\newblock {\em arXiv preprint arXiv:1412.6980}, 2014.

\bibitem{kingma2013auto}
D.~P. Kingma and M.~Welling.
\newblock Auto-encoding variational bayes.
\newblock {\em arXiv preprint arXiv:1312.6114}, 2013.

\bibitem{ledig2016photo}
C.~Ledig, L.~Theis, F.~Husz{\'a}r, J.~Caballero, A.~Cunningham, A.~Acosta,
  A.~Aitken, A.~Tejani, J.~Totz, Z.~Wang, et~al.
\newblock Photo-realistic single image super-resolution using a generative
  adversarial network.
\newblock {\em arXiv preprint arXiv:1609.04802}, 2016.

\bibitem{ma2016group}
K.~Ma, Q.~Wu, Z.~Wang, Z.~Duanmu, H.~Yong, H.~Li, and L.~Zhang.
\newblock Group mad competition-a new methodology to compare objective image
  quality models.
\newblock In {\em Proceedings of the IEEE Conference on Computer Vision and
  Pattern Recognition}, pages 1664--1673, 2016.

\bibitem{mansimov2015generating}
E.~Mansimov, E.~Parisotto, J.~L. Ba, and R.~Salakhutdinov.
\newblock Generating images from captions with attention.
\newblock {\em arXiv preprint arXiv:1511.02793}, 2015.

\bibitem{mirza2014conditional}
M.~Mirza and S.~Osindero.
\newblock Conditional generative adversarial nets.
\newblock {\em arXiv preprint arXiv:1411.1784}, 2014.

\bibitem{nilsback2008automated}
M.-E. Nilsback and A.~Zisserman.
\newblock Automated flower classification over a large number of classes.
\newblock In {\em Computer Vision, Graphics \& Image Processing, 2008.
  ICVGIP'08. Sixth Indian Conference on}, pages 722--729. IEEE, 2008.

\bibitem{odena2016conditional}
A.~Odena, C.~Olah, and J.~Shlens.
\newblock Conditional image synthesis with auxiliary classifier gans.
\newblock {\em arXiv preprint arXiv:1610.09585}, 2016.

\bibitem{radford2015unsupervised}
A.~Radford, L.~Metz, and S.~Chintala.
\newblock Unsupervised representation learning with deep convolutional
  generative adversarial networks.
\newblock {\em arXiv preprint arXiv:1511.06434}, 2015.

\bibitem{reed2016generative}
S.~Reed, Z.~Akata, X.~Yan, L.~Logeswaran, B.~Schiele, and H.~Lee.
\newblock Generative adversarial text to image synthesis.
\newblock In {\em Proceedings of The 33rd International Conference on Machine
  Learning}, volume~3, 2016.

\bibitem{reed2016learning}
S.~E. Reed, Z.~Akata, S.~Mohan, S.~Tenka, B.~Schiele, and H.~Lee.
\newblock Learning what and where to draw.
\newblock In {\em Advances In Neural Information Processing Systems}, pages
  217--225, 2016.

\bibitem{salimans2016improved}
T.~Salimans, I.~Goodfellow, W.~Zaremba, V.~Cheung, A.~Radford, and X.~Chen.
\newblock Improved techniques for training gans.
\newblock In {\em Advances in Neural Information Processing Systems}, pages
  2226--2234, 2016.

\bibitem{shrivastava2016learning}
A.~Shrivastava, T.~Pfister, O.~Tuzel, J.~Susskind, W.~Wang, and R.~Webb.
\newblock Learning from simulated and unsupervised images through adversarial
  training.
\newblock {\em arXiv preprint arXiv:1612.07828}, 2016.

\bibitem{theis2015note}
L.~Theis, A.~v.~d. Oord, and M.~Bethge.
\newblock A note on the evaluation of generative models.
\newblock {\em arXiv preprint arXiv:1511.01844}, 2015.

\bibitem{wang2016generative}
X.~Wang and A.~Gupta.
\newblock Generative image modeling using style and structure adversarial
  networks.
\newblock In {\em European Conference on Computer Vision}, pages 318--335.
  Springer, 2016.

\bibitem{wang2004image}
Z.~Wang, A.~C. Bovik, H.~R. Sheikh, and E.~P. Simoncelli.
\newblock Image quality assessment: from error visibility to structural
  similarity.
\newblock {\em IEEE transactions on image processing}, 13(4):600--612, 2004.

\bibitem{yeh2016semantic}
R.~Yeh, C.~Chen, T.~Y. Lim, M.~Hasegawa-Johnson, and M.~N. Do.
\newblock Semantic image inpainting with perceptual and contextual losses.
\newblock {\em arXiv preprint arXiv:1607.07539}, 2016.

\bibitem{zhang2016stackgan}
H.~Zhang, T.~Xu, H.~Li, S.~Zhang, X.~Huang, X.~Wang, and D.~Metaxas.
\newblock Stackgan: Text to photo-realistic image synthesis with stacked
  generative adversarial networks.
\newblock {\em arXiv preprint arXiv:1612.03242}, 2016.

\end{thebibliography}
}

\end{document}